\documentclass{article}
\usepackage{log_2022}           
\usepackage[numbers,compress,sort]{natbib}

\usepackage{booktabs, tabularx, threeparttable, algorithm, amsmath, amsfonts, mathtools, multirow, multicol, makecell}
\usepackage{enumitem}
\usepackage{listings}

\usepackage{xcolor}

\definecolor{codegreen}{rgb}{0,0.6,0}
\definecolor{codegray}{rgb}{0.5,0.5,0.5}
\definecolor{codepurple}{rgb}{0.54,0,0}
\definecolor{backcolour}{rgb}{0.97,0.97,0.97}

\lstdefinestyle{mystyle}{
    backgroundcolor=\color{backcolour},   
    commentstyle=\color{codegreen},
    keywordstyle=\color{codepurple},
    numberstyle=\tiny\color{codegray},
    stringstyle=\color{codepurple},
    basicstyle=\ttfamily\footnotesize,
    breakatwhitespace=false,         
    breaklines=true,                 
    captionpos=b,                    
    keepspaces=true,                 
    numbers=left,                    
    numbersep=5pt,                  
    showspaces=false,                
    showstringspaces=false,
    showtabs=false,                  
    tabsize=2
}

\title[LCE: An Augmented Combination of Bagging and Boosting in Python]{LCE: An Augmented Combination of Bagging and Boosting\\ in Python}

\author[K. Fauvel et al.]{%
Kevin Fauvel\\
\institute{Inria, Univ Rennes, CNRS, IRISA, France}\\
\email{kevin.fauvel@inria.fr}\And
\'Elisa Fromont\\
\institute{Univ Rennes, IUF, Inria, CNRS, IRISA, France}\\
\email{elisa.fromont@irisa.fr}\And
V\'eronique Masson\\
\institute{Inria, Univ Rennes, CNRS, IRISA, France}\\
\email{veronique.masson@irisa.fr}\And
Philippe Faverdin\\
\institute{PEGASE, INRAE, AGROCAMPUS OUEST, France}\\
\email{philippe.faverdin@inrae.fr}\And
Alexandre Termier\\
\institute{Inria, Univ Rennes, CNRS, IRISA, France}\\
\email{alexandre.termier@inria.fr}
}

\begin{document}
\maketitle

\begin{abstract}
\texttt{lcensemble} is a high-performing, scalable and user-friendly Python package for the general tasks of classification and regression. 
The package implements Local Cascade Ensemble (LCE), a machine learning method that further enhances the prediction performance of the current state-of-the-art methods Random Forest and XGBoost. LCE combines their strengths and adopts a complementary diversification approach to obtain a better generalizing predictor.
The package is compatible with \texttt{scikit-learn}, therefore it can interact with \texttt{scikit-learn} pipelines and model selection tools.
It is distributed under the Apache 2.0 license, and its source code is available at \url{https://github.com/LocalCascadeEnsemble/LCE}.
\end{abstract}

\section{Introduction}
As shown in ``Why Do Tree-Based Models still Outperform Deep Learning on Tabular Data?''~\citep{Grinsztajn22}, the widely used tree-based models remain the state-of-the-art machine learning methods in many cases. Recently, Local Cascade Ensemble (LCE)~\citep{Fauvel19,Fauvel22} proposes to combine the strengths of the top performing tree-based ensemble methods — Random Forest~\citep{Breiman01} and eXtreme Gradient Boosting (XGBoost)~\citep{Chen16}, and integrates a supplementary diversification approach which enables it to be a better generalizing predictor.

LCE has been initially designed for a specific application of classification in~\citep{Fauvel19}, and then used as a building block in~\citep{Fauvel22}. 
However, the implementation of LCE has not yet been published. 
Thus, this article presents \texttt{lcensemble}; a high-performing, scalable and user-friendly Python package implementing LCE for both the classification and regression tasks.
LCE package:
\begin{itemize}[noitemsep,topsep=0pt,parsep=0pt,partopsep=0pt]
    \item Supports parallel processing to ensure scalability;
    \item Handles missing data by design;
    \item Adopts \texttt{scikit-learn}~\citep{scikit-learn} API and provides complete documentation for the ease of use;
    \item Adheres to \texttt{scikit-learn} conventions to allow interaction with \texttt{scikit-learn} pipelines and model selection tools;
    \item Follows software development best practices;
    \item Is released in open source - Apache 2.0 license.
\end{itemize}

\section{LCE Presentation}
The construction of an ensemble method involves combining accurate and diverse individual predictors. There are two complementary ways to generate diverse predictors: (\emph{i}) by changing the training data distribution~\citep{Sharkey97} and (\emph{ii}) by learning different parts of the training data~\citep{Masoudnia14}.

LCE is a method adopting these two diversification approaches. First, (\emph{i}) LCE combines the two well-known methods that modify the distribution of the original training data with complementary effects on the bias-variance trade-off: bagging~\citep{Breiman96} (variance reduction - current reference: Random Forest) and boosting~\citep{Schapire90} (bias reduction - current reference: XGBoost). Then, (\emph{ii}) LCE learns different parts of the training data to capture new relationships that cannot be discovered globally based on a divide-and-conquer strategy (a decision tree).

\begin{figure*}[!htpb]
	\centering
	\includegraphics[width=0.95\linewidth]{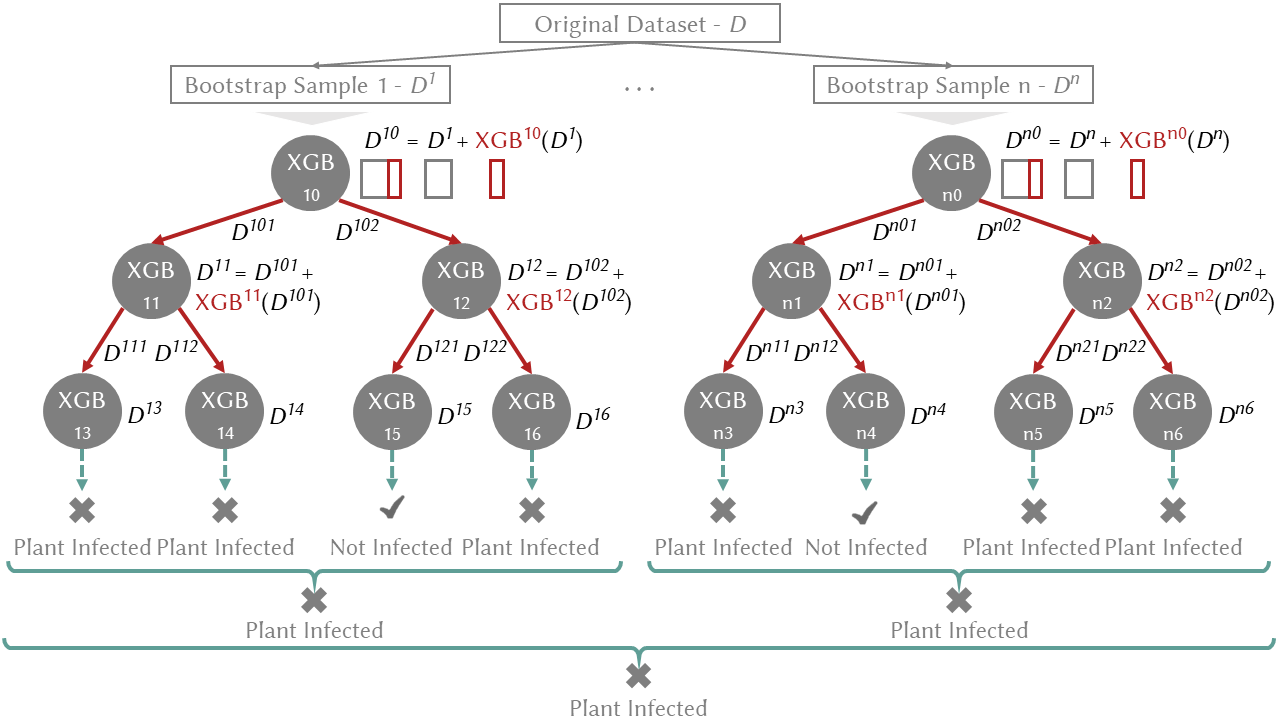}
	\caption{Local Cascade Ensemble on a dataset of plant diseases, with Bagging in cadet blue and Boosting in red. n — number of trees, XGB — XGBoost.}
	\label{fig:lce}
\end{figure*}

Specifically, at each decision node of a tree, a base learner (a boosting algorithm: XGBoost) is applied to the dataset (e.g., see $XGB^{10}$($D^1$) in Figure~\ref{fig:lce}), and its output (e.g., class probabilities for a classifier) are added as new attributes to the dataset in order to propagate boosting down the tree (e.g., see $D^{10}$ in Figure~\ref{fig:lce}). Then, the overfitting generated by the boosted decision tree is mitigated by the use of bagging (see $D^1, ..., D^n$   in Figure~\ref{fig:lce}). Bagging provides variance reduction by creating multiple predictors from random sampling with replacement of the original dataset. 

\section{Package}
\paragraph{Code Quality}
To allow interaction with \texttt{scikit-learn} - the current reference framework for traditional machine learning - and ease the maintenance of the code, LCE package adheres to \texttt{scikit-learn} conventions. Particularly, LCE estimators inherit from \texttt{scikit-learn} base class\footnote{sklearn.base.BaseEstimator} and pass the check estimator\footnote{sklearn.utils.estimator\_checks.check\_estimator}.
Therefore, LCE can interact with \texttt{scikit-learn} pipelines and model selection tools.
Moreover, LCE package follows software development best practices. It integrates continuous integration with the use of CircleCI, extensive unit testing with a code coverage of 100\% for the current package version 0.3.4, and GitHub bug tracker. Tests are executed when commits are made or when a pull request is opened.
 
\paragraph{Dependencies}
The implementation of LCE is compatible with Python 3 ($\geq$ 3.8) and uses some well established Python libraries, without depending on proprietary softwares.
LCE package requires the installation of the hyperparameter optimization library \texttt{hyperopt} \citep{hyperopt}, the scientific computing libraries \texttt{numpy} \citep{numpy} and \texttt{pandas} \citep{pandas}, the machine learning library \texttt{scikit-learn} and \texttt{xgboost}.

\paragraph{Openness to new developers}
LCE is available on GitHub, open source (license Apache 2.0) and compatible with all operating systems.
It includes issue and pull request templates, a contributing file and a code of conduct to encourage the contribution of new developers.
Plus, the code style \texttt{black}, a PEP 8 compliant formatter, is applied in order to ease readability.

\section{Application Programming Interface and Documentation}
\paragraph{API}
The package LCE adopts the same user-friendly API as \texttt{scikit-learn}. It contains two modules: one for the classifier (LCEClassifier) and another one for the regressor (LCERegressor). Each module has a \texttt{fit} method to train the model, and a \texttt{predict} method to make prediction (plus an additional \texttt{predict\_proba} method for LCEClassifier). Once fitted on a training set, a model can be used as predictor on new data. 
We illustrate the interface with the code example below. It trains a LCEClassifier with default parameters on a train set of the public Iris dataset, and computes the accuracy score based on its predictions on the test set.

\begin{lstlisting}[language=Python]
from lce import LCEClassifier
from sklearn.datasets import load_iris
from sklearn.metrics import accuracy_score
from sklearn.model_selection import train_test_split

# Load data and generate a train/test split
data = load_iris()
X_train, X_test, y_train, y_test = train_test_split(data.data, 
                                                    data.target, 
                                                    random_state=0)

# Train LCEClassifier with default parameters
clf = LCEClassifier(n_jobs=-1, random_state=0)
clf.fit(X_train, y_train)

# Make prediction and compute accuracy score
y_pred = clf.predict(X_test)
print("Accuracy: {:.1f}%".format(accuracy_score(y_test, y_pred)*100))
\end{lstlisting}

\begin{lstlisting}[language=Python, caption=LCE classifier on Iris dataset.]
Accuracy: 97.4%
\end{lstlisting}

\paragraph{Documentation}
LCE package is made available with a complete documentation hosted on Read the Docs\footnote{\url{https://lce.readthedocs.io/en/latest/}}.
The documentation includes a presentation of LCE algorithm, the installation procedure on both the Python Package Index (PyPI) and Conda Forge, the full API information about LCE classifier and regressor, and finally a tutorial with simple and non-trivial code examples.
Code examples use public datasets from the library \texttt{scikit-learn} in order to ensure reproducibility.

\section{Evaluation}
In this section, we compare the performance of LCE to that of Random Forest (RF) and XGBoost (XGB) on 10 public datasets randomly selected from the UCI Machine Learning Repository~\citep{Dua17}.
The classical hyperparameters in tree-based learning (max\_depth, n\_estimators) for all models are set by grid search with a 3-fold cross-validation on the training set.
Table~\ref{tab:results} presents the results obtained with the best accuracy for each dataset denoted in boldface.

\begin{table*}[!htpb]
	\caption{Accuracy results on 10 datasets from the UCI repository~\citep{Dua17}. Datasets are sorted in ascending order of their sizes.}
	\label{tab:results}
	\centering
	\scriptsize
	\begin{tabularx}{\linewidth}{m{4cm}>{\centering}m{1.2cm}>{\centering}m{1.2cm}>{\centering}m{1.2cm}|>{\centering}m{1.1cm}>{\centering}m{1.1cm}>{\centering\arraybackslash}m{1.1cm}}
		\toprule
		\textbf{Datasets} & \multicolumn{3}{c|}{\textbf{Description}} & \multicolumn{3}{c}{\textbf{Accuracy (\%)}}\\
        & \textbf{Samples} & \textbf{Dimensions} & \textbf{Classes} & \textbf{RF} & \textbf{XGB} & \textbf{LCE} \\
		\midrule
		Iris & 150 & 4 & 3 & \textbf{97.4} & \textbf{97.4} & \textbf{97.4} \\
        Wine & 178 & 13 & 3 & \textbf{97.8} & 95.6 & \textbf{97.8} \\
        Breast Cancer & 569 & 30 & 2 & 97.2 & 97.2 & \textbf{98.6} \\
        Steel Plates Faults & 1,941 & 27 & 7 & 76.1 & 78.6 & \textbf{79.0} \\
        Wireless Indoor Localization & 2,000 & 7 & 4 & \textbf{97.6} & 96.6 & \textbf{97.6} \\
        Shill Bidding & 6,321 & 11 & 2 & 98.9 & \textbf{99.8} & \textbf{99.8} \\
        Shoppers Purchasing Intention & 12,330 & 17 & 2 & 89.3 & 88.8 & \textbf{89.4}\\
        Nursery & 12,960 & 8 & 5 & 98.7 & \textbf{100.0} & \textbf{100.0} \\
        MAGIC Gamma Telescope & 19,020 & 10 & 2 & 88.4 & 88.1 & \textbf{88.6} \\
        Avila & 20,867 & 10 & 12 & 99.3 & \textbf{99.8} & \textbf{99.8} \\
        \hline
        Average Rank & - & - & - & 2.1 & 2.0 & 1.0 \\
        Number of Wins / Ties & - & - & - & 3 & 4 & 10 \\
    \bottomrule
	\end{tabularx}
\end{table*}

First, we observe that LCE is a better generalizing predictor as it obtains the best average rank across all datasets (LCE: 1.0, Random Forest: 2.1, XGBoost: 2.0).
Then, we can see that combining the strengths of Random Forest and XGBoost, while adopting a supplementary diversification approach, LCE outperforms both methods on 4 out of the 10 datasets. For the rest of the datasets, LCE obtains the same prediction performance as the best performing method between Random Forest and XGBoost (3 wins/ties with Random Forest and 4 wins/ties with XGBoost). Therefore, LCE design allows it to keep the best of both Random Forest and XGBoost methods across the datasets tested, and its supplementary diversification approach can enable it to outperform both of the methods.

\section{Conclusion}
This paper presents \texttt{lcensemble}; a high performing, scalable and user-friendly Python package implementing Local Cascade Ensemble (LCE). LCE is an ensemble method which combines the strengths of the current state-of-the-art methods (Random Forest and XGBoost), and adopts a complementary diversification approach to obtain a better generalizing predictor.
For future work, we plan to render LCE more generic and add the selection of the base learner as a hyperparameter. Therefore, practitioners could replace XGBoost by the method of their choice, which would allow LCE to update with the newest methods or extend its applicability with other existing methods.

\section*{Acknowledgements}
This work was supported by the French National Research Agency under the Investments for the Future Program (ANR-16-CONV-0004).

\bibliographystyle{unsrtnat}
\bibliography{references}

\end{document}